\documentclass[10pt,twocolumn,letterpaper]{article}

\usepackage{cvpr}              %

\usepackage[T1]{fontenc}
\usepackage{tabularx}
\usepackage{array}
\usepackage{makecell}
\usepackage{multirow}
\usepackage{cuted}
\definecolor{cvprblue}{rgb}{0.21,0.49,0.74}
\usepackage[pagebackref,breaklinks,colorlinks,allcolors=cvprblue]{hyperref}
\def\confName{CVPR}
\def\confYear{2027}

\title{TwinScaler: Scaling Up Real-world Robot Demonstrations via \\ Automated Digital Twin Reconstruction, Identification, and Simulation}
\author{%
Wenbo Wang\textsuperscript{1}\quad
Fangyun Wei\textsuperscript{2}\quad
Siyu Xu\textsuperscript{1}\quad
Chenghan Lyu\textsuperscript{1}\quad
Shengjie Wang\textsuperscript{1}\\[2pt]
Jianyi Wang\textsuperscript{1}\quad
Jinjing Zhao\textsuperscript{1}\quad
Yunke Wang\textsuperscript{1}\quad
Jiaolong Yang\textsuperscript{2}\quad
Baining Guo\textsuperscript{2}\quad
Chang Xu\textsuperscript{1}\\[5pt]
\textsuperscript{1}University of Sydney\qquad
\textsuperscript{2}Microsoft Research Asia
}

\makeatletter
\patchcmd{\@maketitle}{\vskip .375in}{\vskip .08in}
  {}{\PackageError{TwinScaler}{Unable to adjust title top spacing}{}}
\patchcmd{\@maketitle}{\vspace*{24pt}}{\vspace*{6pt}}
  {}{\PackageError{TwinScaler}{Unable to adjust title-author spacing}{}}
\patchcmd{\@maketitle}{\vspace*{12pt}}{\vspace*{-6pt}}
  {}{\PackageError{TwinScaler}{Unable to adjust author-teaser spacing}{}}
\makeatother

\begin{document}
\maketitle

\begin{strip}
  \begin{minipage}{\textwidth}
    \centering
    \includegraphics[width=0.99\linewidth]{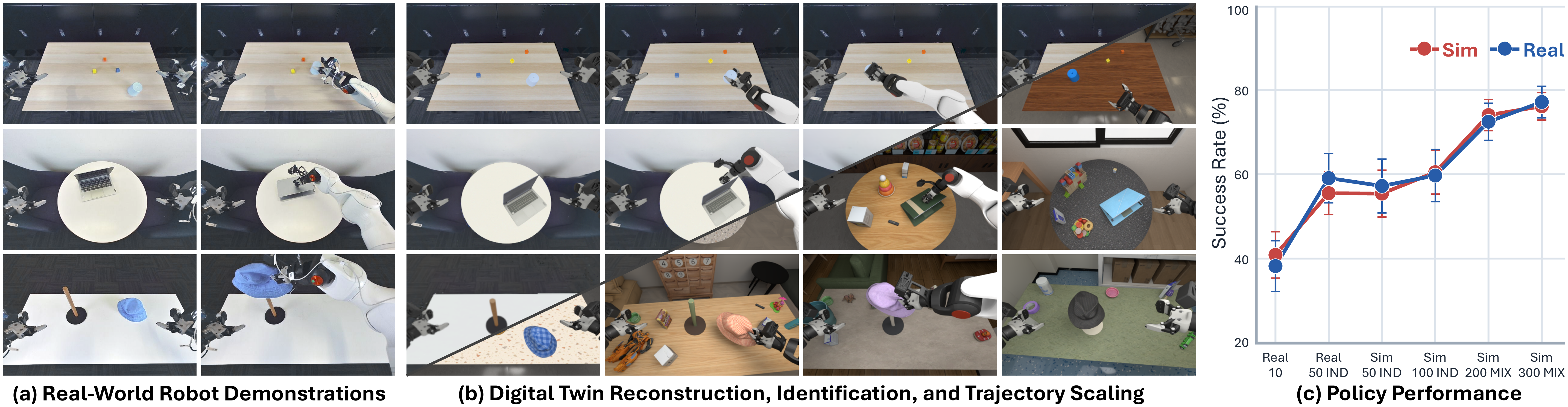}
    \par
    \vspace{-3mm}
    \captionof{figure}{
    From (a) real-world robot demonstrations, TwinScaler automates (b) digital twin reconstruction, identification, and trajectory scaling.
    (c) Scaling 10 real demonstrations to 300 synthetic trajectories per task, with mixed in-distribution (IND) and out-of-distribution (OOD) data, improves mean policy success rates in simulation (red) and the real world (blue), with strong sim-real correlation.
    }
    \label{fig:teaser}
    \vspace{\dimexpr1mm+6pt\relax}
  \end{minipage}
\end{strip}

\begingroup
\setlength{\parskip}{0pt}
\patchcmd{\abstract}
  {\vspace*{12pt}}
  {\vspace*{2pt}}
  {}
  {\PackageError{TwinScaler}{Unable to adjust abstract spacing}{}}
\renewcommand{\endabstract}{}
\begin{abstract}
Robot learning requires scalable robot data; however, real-world data collection is costly and simulation data suffer from sim-real gaps in visual appearance, spatial geometry, interaction outcomes, and physical properties. 
We propose TwinScaler, an automated framework that narrows these four gaps through digital twin reconstruction and identification from real-world robot demonstrations, enabling scalable synthetic data generation without additional scans.
First, a VLM agent provides scene and task understanding with initial object physical priors. Then, iterative texture, scale, and pose optimization aligns the reconstructed assets with real observations, while replay-based physics identification further calibrates object physical parameters using real interaction outcomes and tracked object masks and poses.
Supporting rigid, articulated, and deformable objects, our digital twins enable automated initial-state sampling, trajectory generation, and scene augmentation across in-distribution and out-of-distribution conditions.
Experiments demonstrate correlated policy performance between our digital twins and the real world, with scaled synthetic data providing comparable learning benefits to additional real-world demonstrations. 
Specifically, scaling 10 real-world robot demonstrations to 300 synthetic trajectories per task yields absolute success-rate gains of 35.39\% in simulation and 39.06\% in the real world across in-distribution and out-of-distribution evaluation settings.
Project website: \href{https://dexhand.github.io/TwinScaler_Official/}{https://dexhand.github.io/TwinScaler\_Official/}.

\end{abstract}
\endgroup
\par\newpage

\section{Introduction}
\vspace{-2.5mm}

\begin{table*}[t]
\centering
\rmfamily\fontsize{8}{9}\selectfont
\setlength{\tabcolsep}{1.2pt}
\setlength{\arrayrulewidth}{0.35pt}
\setlength{\aboverulesep}{0pt}
\setlength{\belowrulesep}{0pt}
\setlength{\abovetopsep}{0pt}
\setlength{\belowbottomsep}{0pt}
\setlength{\cmidrulesep}{0pt}
\setlength{\heavyrulewidth}{0.8pt}
\setlength{\lightrulewidth}{0.35pt}
\setlength{\cmidrulewidth}{0.35pt}
\renewcommand{\arraystretch}{1.1111}
\setlength{\extrarowheight}{0pt}
\resizebox{\textwidth}{!}{%
\begin{tabular*}{574.51296pt}{@{\extracolsep{\fill}}clccccccccccc@{}}
\toprule
\multirow{2}{*}{\textbf{Input}} & \multicolumn{1}{c}{\multirow{2}{*}{\textbf{Method}}} & \multicolumn{2}{c}{\textbf{Reconstruction}} & \multicolumn{3}{c}{\textbf{Object Support}} & \multicolumn{2}{c}{\textbf{Real-to-Sim}} & \multicolumn{2}{c}{\textbf{Simulation}} & \multicolumn{2}{c}{\textbf{Sim-to-Real}} \\
\cmidrule(lr){3-4}\cmidrule(lr){5-7}\cmidrule(lr){8-9}\cmidrule(lr){10-11}\cmidrule(l){12-13}
\noalign{\vskip-\cmidrulewidth}
& & \textbf{\makecell[c]{No Extra\\Capture}} & \textbf{\makecell[c]{Include\\Background}} & \textbf{\makecell[c]{Rigid}} & \textbf{\makecell[c]{Articulated}} & \textbf{\makecell[c]{Deformable}} & \textbf{\makecell[c]{Interaction\\Replay}} & \textbf{\makecell[c]{Physics\\Identification}} & \textbf{\makecell[c]{Scene\\Augmentation}} & \textbf{\makecell[c]{Data\\Scaling}} & \textbf{\makecell[c]{Policy\\Transfer}} & \textbf{\makecell[c]{Correlation\\Check}} \\
\midrule
\multirow{7}{*}{\makecell[c]{Scene\\Obs.}}
 & Re3Sim~\cite{Re3Sim} &  & \(\checkmark\) & \(\checkmark\) &  &  &  &  & \(\checkmark\) & \(\checkmark\) & \(\checkmark\) & \(\checkmark\) \\
 & DREAM~\cite{DREAM} &  & \(\checkmark\) & \(\checkmark\) &  &  &  &  & \(\checkmark\) & \(\checkmark\) & \(\checkmark\) & \(\checkmark\) \\
 & EmbodiRSI~\cite{EmbodiRSI} & \(\checkmark\) &  & \(\checkmark\) &  &  &  &  & \(\checkmark\) & \(\checkmark\) & \(\checkmark\) & \(\checkmark\) \\
 & ARSTAG~\cite{ARSTAG} & \(\checkmark\) &  & \(\checkmark\) &  &  &  &  & \(\checkmark\) & \(\checkmark\) & \(\checkmark\) &  \\
 & RoLA~\cite{RoLA} & \(\checkmark\) & \(\checkmark\) & \(\checkmark\) &  &  &  &  & \(\checkmark\) & \(\checkmark\) & \(\checkmark\) &  \\
 & SimFoundry~\cite{SimFoundry} & \(\checkmark\) & \(\checkmark\) & \(\checkmark\) & \(\checkmark\) &  &  &  & \(\checkmark\) & \(\checkmark\) & \(\checkmark\) & \(\checkmark\) \\
 & Genie Sim 3.0~\cite{GenieSim} &  & \(\checkmark\) & \(\checkmark\) & \(\checkmark\) &  &  &  & \(\checkmark\) & \(\checkmark\) & \(\checkmark\) & \(\checkmark\) \\
\midrule
\multirow{6}{*}{\makecell[c]{Human\\Demos}}
 & X-Sim~\cite{X-Sim} &  & \(\checkmark\) & \(\checkmark\) &  &  &  &  & \(\checkmark\) & \(\checkmark\) & \(\checkmark\) &  \\
 & Real2Sem2Real~\cite{Real2Sem2Real} &  &  & \(\checkmark\) &  &  &  &  & \(\checkmark\) & \(\checkmark\) & \(\checkmark\) &  \\
 & Real2Gen~\cite{Real2Gen} & \(\checkmark\) &  & \(\checkmark\) &  &  &  &  & \(\checkmark\) & \(\checkmark\) & \(\checkmark\) &  \\
 & Video2Policy~\cite{Video2Policy} & \(\checkmark\) &  & \(\checkmark\) &  &  &  &  & \(\checkmark\) & \(\checkmark\) & \(\checkmark\) &  \\
 & Video2Sim2Real~\cite{Video2Sim2Real} & \(\checkmark\) &  & \(\checkmark\) &  &  & \(\checkmark\) &  & \(\checkmark\) & \(\checkmark\) & \(\checkmark\) & \(\checkmark\) \\
 & PhysTwin~\cite{PhysTwin} & \(\checkmark\) &  &  &  & \(\checkmark\) & \(\checkmark\) & \(\checkmark\) &  &  & \(\checkmark\) &  \\
\midrule
\multirow{9}{*}{\makecell[c]{Robot\\Demos}}
 & DreMa~\cite{DreMa} &  & \(\checkmark\) & \(\checkmark\) &  &  & \(\checkmark\) &  & \(\checkmark\) & \(\checkmark\) & \(\checkmark\) &  \\
 & D-REX~\cite{D-REX} &  & \(\checkmark\) & \(\checkmark\) & \(\checkmark\) &  & \(\checkmark\) & \(\checkmark\) &  &  & \(\checkmark\) &  \\
 & TwinAligner~\cite{TwinAligner} &  & \(\checkmark\) & \(\checkmark\) & \(\checkmark\) &  & \(\checkmark\) & \(\checkmark\) & \(\checkmark\) &  & \(\checkmark\) & \(\checkmark\) \\
 & RoboSnap~\cite{RoboSnap} & \(\checkmark\) & \(\checkmark\) & \(\checkmark\) & \(\checkmark\) &  & \(\checkmark\) &  & \(\checkmark\) & \(\checkmark\) & \(\checkmark\) & \(\checkmark\) \\
 & Agentic Real2Sim~\cite{AgenticReal2Sim} & \(\checkmark\) & \(\checkmark\) & \(\checkmark\) &  & \(\checkmark\) & \(\checkmark\) &  & \(\checkmark\) & \(\checkmark\) &  & \(\checkmark\) \\
 & Real2Gym~\cite{Real2Gym} & \(\checkmark\) & \(\checkmark\) & \(\checkmark\) & \(\checkmark\) &  & \(\checkmark\) &  & \(\checkmark\) &  & \(\checkmark\) &  \\
 & RoboSimStudio~\cite{RoboSimStudio} & \(\checkmark\) & \(\checkmark\) & \(\checkmark\) & \(\checkmark\) & \(\checkmark\) & \(\checkmark\) & \textendash & \(\checkmark\) & \(\checkmark\)  & \(\checkmark\) &  \\
 & WorldMirror~\cite{WorldMirror} & \(\checkmark\) & \(\checkmark\) & \(\checkmark\) & \(\checkmark\) &  & \(\checkmark\) &  &  &  &  &  \\
 & \textbf{TwinScaler (Ours)} & \(\pmb{\checkmark}\) & \(\pmb{\checkmark}\) & \(\pmb{\checkmark}\) & \(\pmb{\checkmark}\) & \(\pmb{\checkmark}\) & \(\pmb{\checkmark}\) & \(\pmb{\checkmark}\) & \(\pmb{\checkmark}\) & \(\pmb{\checkmark}\) & \(\pmb{\checkmark}\) & \(\pmb{\checkmark}\) \\
\bottomrule
\end{tabular*}}
\vspace{-3.5mm}
\caption{
Comparison of real-to-sim-to-real methods, grouped by their input sources.
\emph{Reconstruction} indicates whether additional capture of objects and scenes is required and whether scene backgrounds are included.
\emph{Object Support} indicates supported object categories.
\emph{Real-to-Sim} assesses support for interaction replay and physical parameter identification.
\emph{Simulation} assesses scene augmentation and synthetic data scaling for policy training.
\emph{Sim-to-Real} covers policy transfer and quantitative comparison of matched sim-real policy performance.
For RoboSimStudio, “\textendash” denotes optional agent-guided physics tuning during replay rather than dedicated physics identification.
}
\vspace{-6mm}
\label{tab:related_work}
\end{table*}

Developing robot systems that can generalize across tasks and environments is a central goal of embodied intelligence. 
While foundation models benefit from internet datasets~\cite{CLIP, GPT3, LLaMA}, obtaining large-scale robot datasets that connect observations, actions, and physical interactions remains costly.
Existing efforts to collect robot data draw on three complementary sources:
Real-world robot trajectories~\cite{DROID, OXE, BridgeData} provide direct and embodiment-aligned supervision, but teleoperation requires substantial human effort and thus is hard to scale up;
Human egocentric videos~\cite{Ego4D, EgoDex, EgoMimic} offer more accessible coverage of diverse manipulation behaviors, yet differences in visual observations, body morphology, and action spaces complicate their use for robot learning;
Simulation can address both limitations by generating trajectories using the target robot's embodiment and action space~\cite{UniGraspTransformer, MobileManiBench, RoboTwin, MimicGen}.
However, sim-to-real transfer remains limited by gaps in visual appearance (e.g., textures and rendering), spatial geometry (e.g., scales and poses), interaction outcomes (e.g., replay consistency), and physical properties (e.g., mass and friction).

To narrow these gaps, Real-to-Sim-to-Real pipelines leverage foundation models and physics simulators to reconstruct simulation from real-world observations, generate synthetic robot trajectories, and train policies for deployment back to real robots. 
These pipelines can be broadly grouped by their primary reconstruction input: scene observations, human demonstrations, and robot demonstrations, as summarized in Table~\ref{tab:related_work}. 
Scene-observation-based methods~\cite{Re3Sim, ARSTAG, SimFoundry, RialTo, ACDC} reconstruct digital scenes and objects for downstream robot execution, while human-demonstration-based methods~\cite{X-Sim, Video2Policy, Video2Sim2Real, PhysTwin} additionally exploit human motions to guide simulated robot behavior. 
These inputs support visual and geometric grounding, but do not provide interaction between the target robot and objects.
Robot demonstrations~\cite{RoboSnap, AgenticReal2Sim, Real2Gym, RoboSimStudio, WorldMirror} provide this real-world interaction, allowing recorded robot trajectories to be replayed in simulation and simulated interactions to be compared with the real-world observations. 
Although such replay supports reconstruction validation and calibration, jointly achieving automated reconstruction, physics identification, and trajectory scaling across diverse robots, objects, and tasks remains challenging.

To address this challenge, we introduce TwinScaler, an automated framework for reconstructing simulation-ready digital twins and scaling manipulation data from real-world robot demonstrations.
Given recorded robot trajectories, multi-view videos, robot models and cameras, TwinScaler reconstructs simulation environments supporting rigid, articulated, and deformable objects without additional scans. 
Our iterative texture, scale, and pose optimization improves visual and geometric alignment, while our replay-based physics identification calibrates object physical parameters by matching simulated interactions, object masks, and poses to their real-world counterparts.
The reconstructed and identified digital twins can expand a limited set of real-world robot demonstrations into a scalable source of manipulation data covering diverse object initial states and visual appearances. 
Beyond validating the reconstruction quality and interaction replay consistency, we also evaluate the learning benefits with our scaled synthetic data and assess policy generalization across in-distribution and out-of-distribution conditions in simulation and the real world. All the code, datasets, and models will be publicly released.
\vspace{-6mm}
\section{Related Work}
\label{sec:related_work}
\vspace{-2mm}

\noindent\textbf{Real-to-Sim-to-Real.} 
Simulation enables scalable robot data generation, policy training, and evaluation~\cite{MuJoCo, ManiSkill, IsaacLab, SIMPLER, PolaRiS}, but gaps in visual appearance, spatial geometry, interaction outcomes, and physical properties limit the sim-to-real transfer performance. Real-to-Sim-to-Real pipelines narrow these gaps by reconstructing and grounding simulation environments from real-world observations: scene observations, human demonstrations, and robot demonstrations, as summarized in Table~\ref{tab:related_work}. Scene-observation-based methods~\cite{Re3Sim, ARSTAG, SimFoundry} reconstruct scene assets and layouts before introducing robots and generating trajectories in simulation. Human-demonstration-based methods~\cite{Real2Gen, Video2Policy, Video2Sim2Real} further extract task goals and motion priors from human videos to guide simulated robot behavior. Both approaches reduce visual and geometric gaps by grounding simulation in real-world observations.
However, their inputs do not directly capture the target robot's interactions with objects, limiting supervision for validating simulation interaction outcomes and calibrating object physical properties.

Robot-demonstration-based methods provide the direct supervision of robot–object interactions, enabling recorded robot trajectories to be replayed in simulation and simulated outcomes to be compared with real-world observations. 
D-REX~\cite{D-REX} and TwinAligner~\cite{TwinAligner} exploit these interactions to identify object physical parameters but require additional object scans. 
Recent advances in VLM agents have enabled Agentic Real2Sim~\cite{AgenticReal2Sim}, Real2Gym~\cite{Real2Gym}, RoboSimStudio~\cite{RoboSimStudio}, and WorldMirror~\cite{WorldMirror} to automate reconstruction through tool use without dedicated object scans. These frameworks use interaction feedback to refine imperfect reconstructed scenes and improve replay consistency. 
However, their object physical parameters may still rely on VLM estimates, potentially leaving gaps between simulated and real-world object physical properties.
Therefore, achieving automated digital twin reconstruction with aligned interaction outcomes and object physics remains challenging. To address this, we propose TwinScaler, an automated framework for digital twin reconstruction and physics identification.
Our proposed iterative object texture, scale, and pose optimization improves visual and geometric alignment, providing a consistent foundation for interaction replay-based physics identification to calibrate object physical properties from real-world observations.

\noindent\textbf{Synthetic Data Generation.} 
Collecting robot datasets across diverse initial states and visual conditions is costly, motivating synthetic data scaling through two complementary mechanisms: trajectory generation and scene augmentation. Trajectory generation adapts demonstrated robot trajectories to new object configurations. MimicGen~\cite{MimicGen} replays transformed object-centric trajectory segments and connects them through interpolation, while SkillMimicGen~\cite{SkillMimicGen} adapts contact-rich skill segments and uses motion planning to connect them across larger scene variations. On the other hand, scene augmentation further broadens the visual coverage while preserving the task semantics. X-Sim~\cite{X-Sim} varies camera viewpoints, Real2Gym~\cite{Real2Gym} augments backgrounds and lighting, SimFoundry~\cite{SimFoundry} modifies object textures and material appearance and introduces distractors in simulation. Based on these achievements, TwinScaler automates both mechanisms within reconstructed digital twins, combining initial state sampling, trajectory generation, and scene augmentation to scale up real-world robot demonstrations across in-distribution and out-of-distribution conditions. Experiments demonstrate that our scaled synthetic data provide comparable learning benefits to additional real-world demonstrations. Scaling 10 real-world demonstrations to 300 synthetic trajectories per task yields average absolute success-rate gains of 35.39\% in simulation and 39.06\% in the real world across in-distribution and out-of-distribution evaluation settings.

\vspace{-2mm}
\section{Method}
\label{sec:method}
\vspace{-1mm}

\begin{figure*}[!t]
\centering
\includegraphics[width=\textwidth]{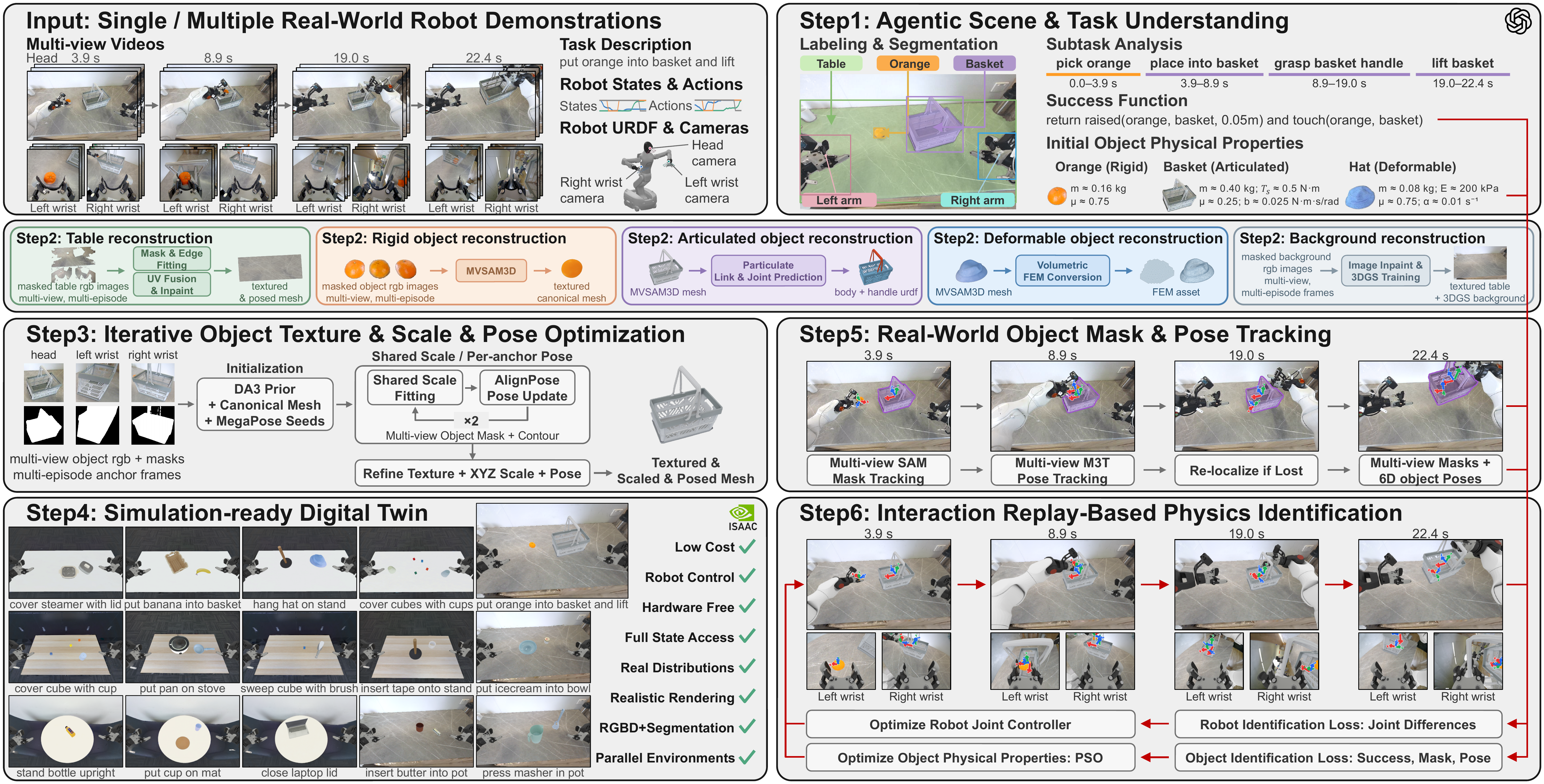}
\vspace{-7mm}
\caption{
Overview of our automated digital twin reconstruction and identification pipeline. 
VLM-based understanding guides asset reconstruction and provides object physical priors. 
Iterative optimization aligns object textures, scales, and poses, while replay-based physics identification calibrates object physics from observed interactions.
Together, these stages transform real-world robot demonstrations into simulation-ready digital twins with sim-real aligned visual appearance, spatial geometry, interaction outcomes, and physical properties.
}
\vspace{-5mm}
\label{fig:twin_framework}
\end{figure*}

\textbf{Overview.}
As shown in Figure~\ref{fig:twin_framework}, given single or multiple real-world robot demonstrations of the same task, each comprising multi-view videos, robot states, and actions, along with a task description, robot models and cameras, TwinScaler automatically reconstructs a simulation-ready digital twin and scales up synthetic data without additional scans.
A VLM agent~\cite{GPT6Sol} first provides scene and task semantic understanding with initial object physical priors (Sec. 3.1). Then, specialized models~\cite{MV-SAM3D, Particulate, FEM, 3DGS} reconstruct the corresponding digital assets, which are refined based on the real-world observations through the iterative texture, scale, and pose optimization (Sec. 3.2). The replay-based physics identification further calibrates object physical parameters using real-world interactions, tracked object masks and poses (Sec. 3.3). Finally, our digital twins support automated initial state sampling, synthetic trajectory generation, and visual appearance augmentation across in-distribution and out-of-distribution conditions (Sec. 3.4). Experiments demonstrate that scaled synthetic data provide comparable learning benefits to additional real-world demonstrations, while policy success rates in our digital twins correlate with those in the real world.

\vspace{-1mm}
\subsection{Agentic Scene and Task Understanding}
\vspace{-1mm}
TwinScaler combines VLM-based semantic understanding with explicit optimization to narrow the sim-real gaps in visual appearance, spatial geometry, interaction outcomes, and physical properties.
As shown in Figure~\ref{fig:twin_framework}, GPT~\cite{GPT6Sol} extracts structured scene and task information from the task descriptions and selected demonstration frames. 
It builds a shared task catalog containing the table shape, object names and categories, identified bounding-boxes and SAM3~\cite{SAM} masks. 
Then, the agent estimates object physical properties, including mass and surface friction for all objects, joint friction and damping for articulated objects, and stiffness and damping for deformable objects. These physical priors initialize the subsequent replay-based physics identification (Sec. 3.3).
Finally, using gripper events, robot states, and sampled images, the agent annotates subtask goals, temporal intervals, associated objects, and a task success function for the subsequent synthetic trajectory scaling (Sec. 3.4).

\vspace{-1mm}
\subsection{Automated Digital Twin Reconstruction}
\vspace{-1mm}
\noindent\textbf{Digital Asset Reconstruction.}
Using the VLM-generated task catalog, TwinScaler reconstructs table, object, and background assets from segmented multi-view observations across single or multiple robot demonstrations.
For the table, TwinScaler fits a shared tabletop plane to multi-view SAM~\cite{SAM} masks and edges, then projects and fuses the observed table RGB values into a UV texture map, filling missing regions through image inpainting~\cite{Qwen}.
For each task object or distractor, TwinScaler uses MV-SAM3D~\cite{MV-SAM3D} to reconstruct a textured canonical mesh from multi-view RGB images and masks.
For articulated objects, TwinScaler uses Particulate~\cite{Particulate} to partition the canonical mesh into links and predict a URDF kinematic tree specifying joint types, axes, and limits.
For deformable objects, TwinScaler converts the canonical mesh into volumetric FEM~\cite{FEM} assets while retaining textured surfaces, with stiffness and damping initialized from VLM priors.
For the background, TwinScaler masks out the table, objects, and robots, inpaints~\cite{Qwen} occluded regions outside the tabletop, and reconstructs the surrounding background using 3D Gaussian Splatting~\cite{3DGS} from multiple frames and views. 

\noindent\textbf{Optimization-based Alignment.}
The reconstructed digital assets may still differ from real-world observations in texture, scale, and pose. Therefore, TwinScaler introduces iterative optimization to reduce these discrepancies using multi-view observations from single or multiple robot demonstrations. For each object and episode, TwinScaler selects the earliest frame in which the object is visible in at least two camera views as the optimization anchor frame. After alignment, TwinScaler further refines the first-frame object poses to start the interaction replay (Sec. 3.3).

First, multi-view DA3~\cite{DA3} initializes the episode-shared object scale, while MegaPose~\cite{MegaPose} provides per-anchor pose priors.
Then, TwinScaler performs two rounds of alternating shared-scale fitting and AlignPose-based~\cite{AlignPose} anchor-pose refinement under tabletop support and penetration constraints, minimizing the following alignment objective:
\vspace{-1.5mm}
\begin{equation}
\begin{aligned}
\mathcal{L}_{\mathrm{align}}
={}&\left(1-\overline{\mathrm{IoU}}\right)
+0.25\left(1-\mathrm{IoU}_{\min}\right)\\
&+0.10\,\mathcal{L}_{\mathrm{contour}}
+\mathcal{L}_{\mathrm{table}},
\end{aligned}
\label{eq:geometric_alignment}
\vspace{-2mm}
\end{equation}
where $\overline{\mathrm{IoU}}$ and $\mathrm{IoU}_{\min}$ denote the mean and minimum silhouette IoU across valid camera views, respectively. $\mathcal{L}_{\mathrm{contour}}$ measures normalized bidirectional contour error, while $\mathcal{L}_{\mathrm{table}}$ encodes tabletop-related geometric penalties.
For articulated and deformable objects, TwinScaler first aligns their canonical meshes before refining per-anchor and first-frame joint configurations and deformation states.

Finally, with geometry and poses aligned and fixed, TwinScaler calibrates object texture colors against the observed RGB values using color gains shared across views and episodes. 
Together, digital asset reconstruction and optimization-based alignment produce a simulation-ready digital twin in Isaac Sim 6.0~\cite{IsaacSim}, establishing a visual and geometric foundation for replay-based physics identification (Sec. 3.3) and synthetic trajectory scaling (Sec. 3.4).

\begin{figure*}[!t]
\centering
\includegraphics[width=\textwidth]{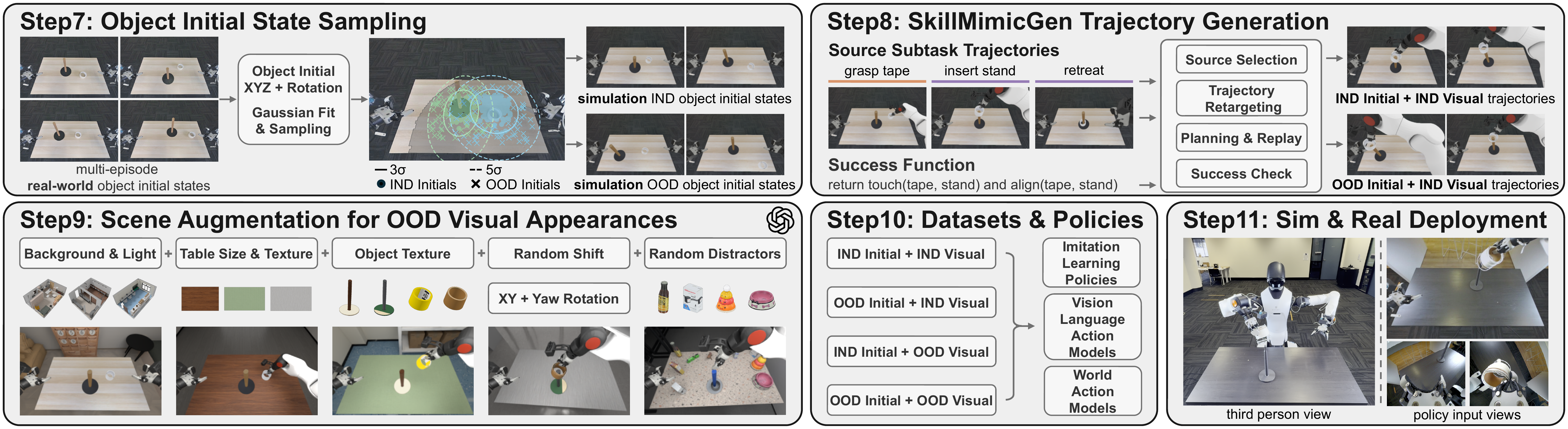}
\vspace{-7mm}
\caption{
Overview of our automated data scaling pipeline. Gaussian sampling expands object initial state coverage, SkillMimicGen generates successful trajectories, and scene augmentation diversifies visual appearances. 
Together, they produce four groups of synthetic trajectories spanning IND/OOD initial states and visual appearances for policy training and deployment in simulation and the real world.
}
\vspace{-5mm}
\label{fig:scaler_framework}
\end{figure*}

\vspace{-1mm}
\subsection{Replay-based Physics Identification}
\vspace{-1mm}
Robot demonstrations provide recorded robot trajectories and object responses, enabling direct comparison between real and simulated interactions.
Prior automated methods, such as the Agentic Real2Sim~\cite{AgenticReal2Sim} and Real2Gym~\cite{Real2Gym}, use replay feedback to refine object placement and improve replay feasibility while relying on VLM-estimated object physical priors.
TwinScaler steps further by identifying the effective object physical parameters through interaction replay. 
However, the geometric and physical errors can compensate for each other during joint optimization, degrading alignment with real-world observations and obscuring which changes improve replay outcomes.
Therefore, TwinScaler fixes the object scales and poses optimized in Section 3.2, preserving their geometric alignment while optimizing the object physical parameters through interaction replay.

First, TwinScaler calibrates the simulated robot controller by matching real-world and simulated joint states under the recorded robot actions.
Then, TwinScaler obtains real-world object masks and poses through multi-view SAM~\cite{SAM} and M3T~\cite{M3T} tracking, using MegaPose~\cite{MegaPose} and AlignPose~\cite{AlignPose} to recover from tracking failures. 
For articulated and deformable objects, TwinScaler further refines their per-frame joint configurations and deformation states.
Starting from the VLM-estimated physical priors, particle swarm optimization (PSO)~\cite{PSO} searches for object physical parameters shared across robot demonstrations. These include mass and surface friction for all task objects, joint friction and damping for articulated objects, and stiffness and damping for deformable objects. 
Each candidate parameter set is evaluated by replaying the recorded robot actions from the first frame under simulated dynamics, with PSO minimizing the interaction objective:
\vspace{-1mm}
\begin{equation}
\mathcal{L}_{\mathrm{interact}}
=
\left(1-\mathrm{SR}_{\mathrm{replay}}\right)
+
0.5\,\mathcal{L}_{\mathrm{mask}}
+
0.5\,\mathcal{L}_{\mathrm{pose}},
\label{eq:physics_identification}
\vspace{-1mm}
\end{equation}
where $\mathrm{SR}_{\mathrm{replay}}$ denotes the replay success rate in $[0,1]$.
$\mathcal{L}_{\mathrm{mask}}$ measures the quality-weighted mean silhouette mismatch between SAM and simulated object masks during replay, while $\mathcal{L}_{\mathrm{pose}}$ normalizes the quality-weighted mean position error between tracked and simulated object poses. Both terms are averaged across valid episodes and task objects.
Finally, TwinScaler incorporates the identified object physical parameters into the digital twins, improving sim-real interaction and physics consistency while preserving visual and geometric alignment.

\vspace{-1mm}
\subsection{Synthetic Trajectory Scaling}
\vspace{-1mm}
Using the reconstructed and identified digital twins, TwinScaler scales robot demonstrations across in-distribution and out-of-distribution object initial states and scene visual appearances through three stages, as shown in Figure~\ref{fig:scaler_framework}.

\noindent\textbf{Object Initial State Sampling.}
First, TwinScaler fits per-object Gaussian models to their initial positions and symmetry-aware orientations across all demonstration episodes. 
It defines the region within $3\sigma$ as in-distribution (IND) and the region beyond $3\sigma$ and within $5\sigma$ as out-of-distribution (OOD).
TwinScaler then samples IND and OOD object initial states, requiring all objects to lie within their respective IND regions for IND sampling and at least one object to lie within its OOD region for OOD sampling.

\noindent\textbf{SkillMimicGen Trajectory Generation.}
Then, TwinScaler uses SkillMimicGen~\cite{SkillMimicGen} to scale up successful replays of real-world robot demonstrations.
Guided by VLM-annotated subtask goals, intervals, and reference objects, SkillMimicGen organizes demonstrated robot trajectories into object-referenced skill segments.
For each sampled IND or OOD object initial states, it selects a compatible source demonstration and retargets the skill trajectories according to the current reference-object states.
Motion planning~\cite{cuRobo} connects the retargeted skill segments, which are then directly replayed with their associated gripper commands. Only trajectories satisfying the VLM-derived task success function are retained.

\noindent\textbf{Scene Visual Augmentation.}
Finally, TwinScaler creates out-of-distribution visual appearances through VLM texture generation and procedural scene augmentation.
As shown in Figure~\ref{fig:scaler_framework}, TwinScaler varies backgrounds and table dimensions, introduces VLM-generated~\cite{GPT6Sol} table and object textures, randomizes the relative translation and rotation between robot-object and their surrounding scene, and places tabletop distractors~\cite{GSO}.
Together, these scene visual augmentations expand each successful manipulation trajectory into visually diverse training data while preserving the underlying task-object geometry and robot-object interactions.

\begin{figure*}[!t]
  \centering
  \includegraphics[width=\textwidth]{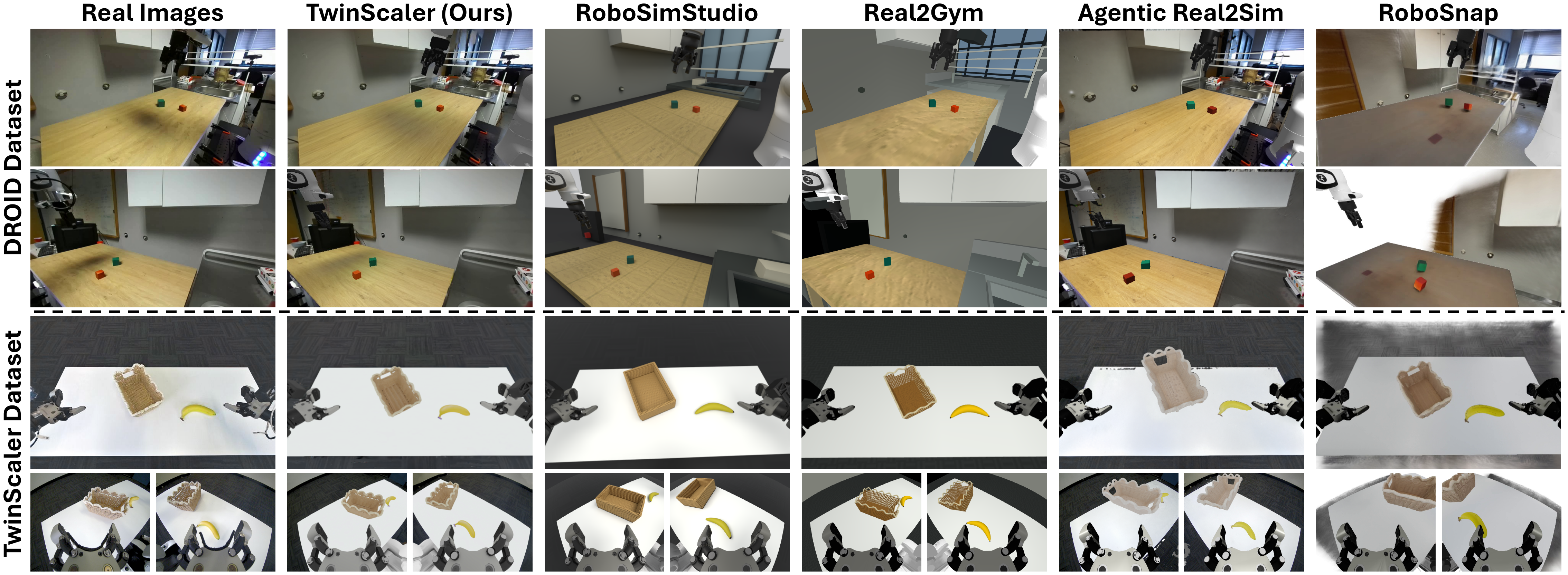}
  \vspace{-7mm}
  \caption{Qualitative comparison of digital twins reconstructed by different methods on the TwinScaler and DROID datasets, highlighting differences in visual appearance and spatial geometry across multiple viewpoints. 
  }
  \vspace{-2.5mm}
  \label{fig:baseline_qualitative}
\end{figure*}

\begin{table*}[!t]
\centering
\rmfamily\fontsize{8.5}{10}\selectfont
\setlength{\tabcolsep}{3pt}
\renewcommand{\arraystretch}{1.15}
\begin{tabular*}{\textwidth}{@{\extracolsep{\fill}}l*{7}{c}@{}}
\toprule
\multirow{2}{*}{\textbf{Method}}
& \multicolumn{2}{c}{\textbf{Interaction Quality}}
& \multicolumn{3}{c}{\textbf{Visual Quality}}
& \multicolumn{2}{c}{\textbf{Efficiency}} \\
\cmidrule(lr){2-3}\cmidrule(lr){4-6}\cmidrule(l){7-8}
& \mbox{\textbf{Replay Success}$\uparrow$}
& \mbox{\textbf{Object Mask IoU}$\uparrow$}
& \mbox{\textbf{PSNR (dB)}$\uparrow$}
& \mbox{\textbf{SSIM}$\uparrow$}
& \mbox{\textbf{LPIPS}$\downarrow$}
& \mbox{\textbf{Time (h)}$\downarrow$}
& \mbox{\textbf{Token Cost (\$)}$\downarrow$} \\
\midrule
RoboSnap & \mbox{31.25\% / 28.13\%} & \mbox{0.166 / 0.179} & \mbox{10.83 / 11.49} & \mbox{0.483 / 0.506} & \mbox{0.645 / 0.656} & 0.56 & \textbf{0.23} \\
Agentic Real2Sim & \mbox{46.88\% / 53.13\%} & \mbox{0.282 / 0.318} & \mbox{12.45 / \underline{15.54}} & \mbox{0.457 / \underline{0.586}} & \mbox{0.538 / \underline{0.383}} & \textbf{0.42} & 0.63 \\
Real2Gym & \mbox{56.25\% / 62.50\%} & \mbox{\underline{0.513} / \underline{0.556}} & \mbox{\underline{13.58} / 12.83} & \mbox{0.619 / 0.558} & \mbox{\underline{0.419} / 0.489} & 0.95 & 1.28 \\
RoboSimStudio & \mbox{\underline{65.63\%} / \underline{71.88\%}} & \mbox{0.393 / 0.458} & \mbox{13.13 / 13.56} & \mbox{\underline{0.625} / 0.552} & \mbox{0.475 / 0.502} & 0.82 & 1.67 \\
\textbf{TwinScaler (Ours)} & \mbox{\textbf{93.75\%} / \textbf{81.25\%}} & \mbox{\textbf{0.672} / \textbf{0.635}} & \mbox{\textbf{16.15} / \textbf{17.34}} & \mbox{\textbf{0.707} / \textbf{0.727}} & \mbox{\textbf{0.328} / \textbf{0.282}} & \underline{0.48} & \underline{0.32} \\
\bottomrule
\end{tabular*}
\vspace{-2.5mm}
\caption{Quantitative comparison of interaction quality, visual quality, and reconstruction efficiency across different real-to-sim methods. Interaction and visual quality are evaluated by replaying recorded robot actions in reconstructed scenes. 
Paired values report TwinScaler / DROID results, each averaged over 32 selected episodes. Bold and underlined values indicate the best and second-best results, respectively.}
\vspace{-5.5mm}
\label{tab:baseline_quantitative}
\end{table*}

\vspace{-1mm}
\subsection{Policy Training and Deployment}
\vspace{-1mm}
Scaling trajectories across IND and OOD initial states and visual appearances yields four groups of synthetic data.
Alongside multi-view RGB images, robot states and actions, TwinScaler can also provide depth maps, segmentation masks, object states, and contact annotations.
These multimodal data support training diverse policies and models~\cite{BC, DiffusionPolicy, OpenVLA, pi05, WorldVLA, DreamZero} for deployment under IND and OOD conditions in simulation and the real world (Sec. 4.2).

\vspace{-2mm}
\section{Experiments}
\label{sec:experiments}
\vspace{-1mm}

\noindent\textbf{TwinScaler Dataset.} 
We first collect the TwinScaler dataset through human teleoperation~\cite{MetaQuest} of a Genie G2 robot~\cite{GenieG2}, comprising 160 real-world robot demonstrations across 16 manipulation tasks in five tabletop scenes.
The dataset covers rigid, articulated, and deformable objects, with diverse single-arm and dual-arm tasks including pick-and-place, insertion, covering, lifting, sweeping, closing, hanging and bottle uprighting.
Each demonstration includes multi-view videos, robot states and actions, task descriptions, robot models and cameras.
This dataset supports evaluating our digital twin reconstruction, replay-based identification, and downstream trajectory scaling for policy training and deployment.
Figure~\ref{fig:twin_framework} shows simulation-ready digital twins reconstructed from the TwinScaler dataset.

\noindent\textbf{Experiment Overview.}
We then compare TwinScaler with real-to-sim baseline methods in terms of interaction quality, visual quality, and reconstruction efficiency on selected demonstrations from the TwinScaler and DROID~\cite{DROID} datasets (Sec. 4.1). We also evaluate synthetic trajectory scaling and policy performance on eight TwinScaler tasks across four IND and OOD conditions in simulation and the real world (Sec. 4.2), perform the sim-real correlation check (Sec. 4.3), and conduct the ablation studies (Sec. 4.4).

\begin{figure*}[!t]
  \centering
  \includegraphics[width=\textwidth]{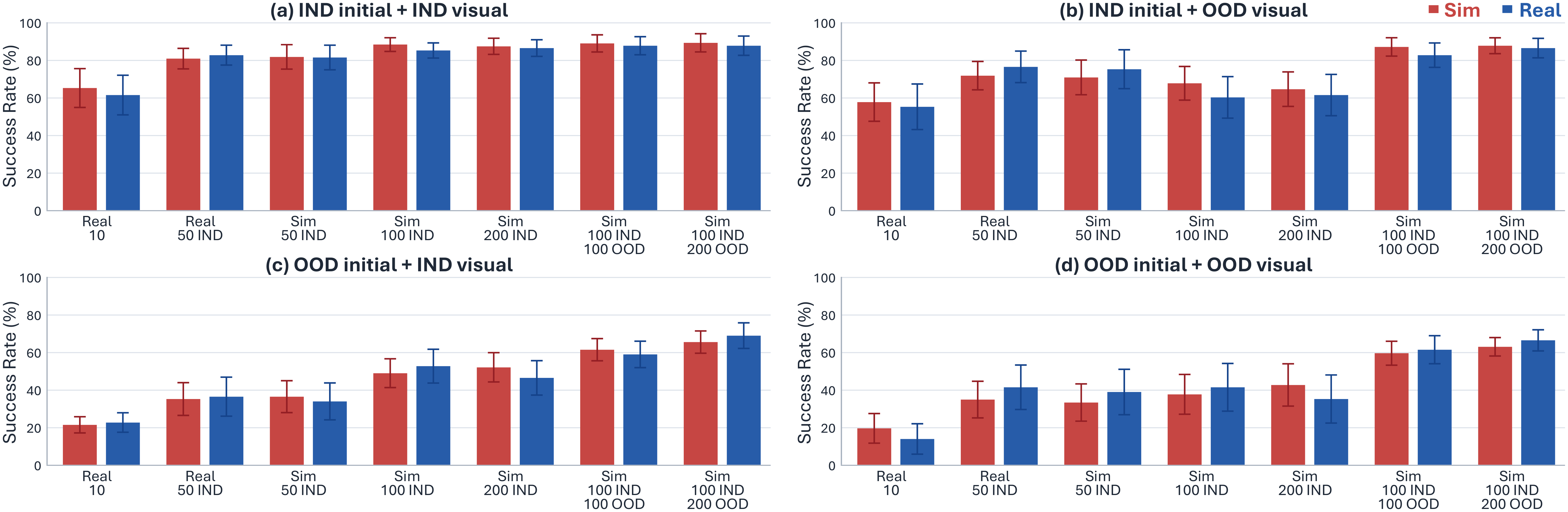}
  \vspace{-7mm}
  \caption{Policy success rates with increasing amounts of training data, evaluated in simulation (red) and the real world (blue) under four IND/OOD conditions. Filled bars show mean success rates, while error bars indicate $\pm1$ standard error across eight TwinScaler tasks.}
  \vspace{-6mm}
  \label{fig:inference_quantitative}
\end{figure*}

\vspace{-1mm}
\subsection{Real-to-Sim Baseline Comparison}
\label{sec:real2sim_baseline}
\vspace{-1mm}

We compare TwinScaler with RoboSnap~\cite{RoboSnap}, Agentic Real2Sim~\cite{AgenticReal2Sim}, Real2Gym~\cite{Real2Gym}, and RoboSimStudio~\cite{RoboSimStudio} on 32 selected demonstrations from each of the TwinScaler and DROID datasets. 
We evaluate the reconstructed scenes during interaction replay, using replay success rates and object mask IoUs to assess interaction quality, and PSNR, SSIM, and LPIPS to assess visual quality. We also measure the reconstruction efficiency through per-episode runtime and agent token cost, with all methods using GPT-6.1 Sol~\cite{GPT6Sol}.

RoboSnap~\cite{RoboSnap} and Agentic Real2Sim~\cite{AgenticReal2Sim} rely on single-view reconstruction, with discrepancies in object scales, poses, and visual appearance across viewpoints, as shown in Figure~\ref{fig:baseline_qualitative}. 
Real2Gym~\cite{Real2Gym} uses multi-view observations and agent-guided replay correction, achieving better alignment and higher replay success, while visual and geometric discrepancies remain in its simplified object and scene assets.
RoboSimStudio~\cite{RoboSimStudio} incurs the highest token cost, with VLM agents handling both scene reconstruction and replay-based refinement, including object physics tuning when needed. 
Although this agentic workflow achieves the highest replay success among the baselines, its simplified assets retain visual and geometric discrepancies, as reflected in its object mask IoU and visual quality scores in Table~\ref{tab:baseline_quantitative}.

TwinScaler combines the multi-view optimization-based alignment with dedicated replay-based physics identification.
Without agentic refinement, it achieves direct replay success rates of 93.75\%/81.25\% and object mask IoUs of 0.672/0.635 on the TwinScaler/DROID datasets, alongside the best visual quality scores and competitive efficiency.

\vspace{-2.5mm}
\subsection{Trajectory Scaling for Generalizable Policies}
\label{sec:scaling_policy}
\vspace{-1.5mm}

\noindent\textbf{TwinScaler Benchmark.} 
We evaluate eight TwinScaler tasks across four combinations of IND and OOD initial states and visual appearances. 
For each task in simulation, we sample 40 + 40 IND and 40 + 40 OOD initial states for evaluation across the four conditions.
Each simulated task uses one reconstructed IND visual appearance and six augmented OOD visual appearances, with four used for training and two held out for evaluation.
For real-world evaluation, we use one IND and one OOD visual appearance per task and sample 10 initial states per evaluation condition using the corresponding IND/OOD sampling procedure.

\noindent\textbf{Trajectory Scaling.}
From 10 real-world demonstrations per task, TwinScaler uses SkillMimicGen and scene augmentation to generate 200 IND and 200 OOD synthetic trajectories.
The IND trajectories use IND initial states and visual appearance, while the OOD trajectories combine IND initial + OOD visual, OOD initial + IND visual, and OOD initial + OOD visual conditions in a 1:2:2 ratio. 
Half of the OOD visual training and evaluation samples include random shifts and distractors. 
To compare learning benefits, we additionally collect 50 real IND demonstrations per task.

\noindent\textbf{Policy Performance.}
To evaluate the benefits of trajectory scaling, we train one $\pi_{0.5}$~\cite{pi05} policy for 30,000 steps on each of the seven training sets. Each set includes trajectories from all eight tasks. Figure~\ref{fig:inference_quantitative} reveals the following findings:

\emph{(1) Synthetic data provide comparable learning benefits to additional real-world demonstrations.}
Sim 50 IND achieves an average real-world success rate of 57.19\%, compared with 59.06\% for Real 50 IND, with differences of at most 2.50\% across the four evaluation conditions.

\emph{(2) Simulated and real-world policy success rates are correlated.}
Across the seven training configurations and four evaluation conditions, simulated and real-world success rates exhibit similar trends, with a mean absolute error of 3.40\%. We quantify this correlation in Section 4.3.

\emph{(3) Initial state generalization is more challenging than visual appearance generalization.}
For Sim 50 IND, simulated success rate decreases from 81.56\% to 70.63\% when only visual appearances become OOD, but drops to 36.25\% when only object initial states become OOD.
Real-world evaluation shows the same pattern.

\emph{(4) Increasing IND data alone does not consistently improve OOD generalization.}
Scaling IND training data alone from 50 to 200 trajectories improves performance under IND initial + IND visual, but reduces success rates under IND initial + OOD visual from 70.63\% to 64.38\% in simulation and from 75.00\% to 61.25\% in the real world.

\emph{(5) Introducing OOD data improves both generalization.}
Compared with Sim 100 IND, adding 100 OOD trajectories raises real-world success rates from 51.25\% to 67.50\% across three OOD evaluation conditions. 
With 200 trajectories per task, mixed IND/OOD training yields higher mean success rates than IND-only training in simulation (74.06\% vs.\ 61.48\%) and in the real world (72.50\% vs.\ 57.19\%), demonstrating the benefits of broader training coverage.

\emph{(6) Trajectory scaling improves policy performance.}
Scaling 10 real-world demonstrations to 300 IND and OOD synthetic trajectories per task raises mean success rates from 40.78\% to 76.17\% in simulation and from 38.13\% to 77.19\% in the real world across four evaluation conditions.

\noindent\textbf{Baseline Comparison.} 
We compare digital twins reconstructed by TwinScaler, RoboSimStudio, Real2Gym, and Agentic Real2Sim for trajectory generation and policy training, using the same eight TwinScaler tasks, each with 10 real demonstrations.
From successful replays in each baseline's reconstructed digital twins, we generate 100 IND and 200 OOD synthetic trajectories per task using the same protocol (Sec. 3.4).
For each method, we train a $\pi_{0.5}$ policy across all eight tasks for 30,000 steps and evaluate it under the same four IND/OOD conditions in that method's reconstructed twin and the real world.
As shown in Table~\ref{tab:policy_baseline_comparison}, with matched training set size, TwinScaler achieves the highest real-world policy success rate and sim-real Pearson correlation among all methods. 

\begin{table}[t]
  \centering
  \footnotesize
  \setlength{\tabcolsep}{1pt}
  \begin{tabular*}{\columnwidth}{@{\extracolsep{\fill}}lccc@{}}
    \toprule
    \multirow{2}{*}{\textbf{Method}} &
    \textbf{Simulation} & \textbf{Real-World} & \multicolumn{1}{c@{}}{\textbf{Sim-Real}} \\
    \cmidrule(lr){2-2}\cmidrule(lr){3-3}\cmidrule(l){4-4}
    & \textbf{Policy SR}$\uparrow$ & \textbf{Policy SR}$\uparrow$ & \textbf{Correlation} \textit{r}\,$\uparrow$ \\
    \midrule
    Agentic Real2Sim & 46.02\% & 51.56\% & 0.911 \\
    Real2Gym & 48.83\% & 54.69\% & 0.905 \\
    RoboSimStudio & 57.34\% & 62.19\% & 0.923 \\
    \textbf{TwinScaler (Ours)} & \textbf{76.17\%} & \textbf{77.19\%} & \textbf{0.976} \\
    \bottomrule
  \end{tabular*}
  \vspace{-3.5mm}
  \caption{
  Comparison of policy performance using 300 trajectories per task, generated from each method's digital twins. 
  Success rates are averaged over eight tasks and four IND/OOD evaluations. 
  Pearson $r$ is computed from these 32 paired sim-real success rates.
  }
  \vspace{-2mm}
  \label{tab:policy_baseline_comparison}
\end{table}

\begin{figure}[t]
  \centering
  \includegraphics[width=0.98\columnwidth]{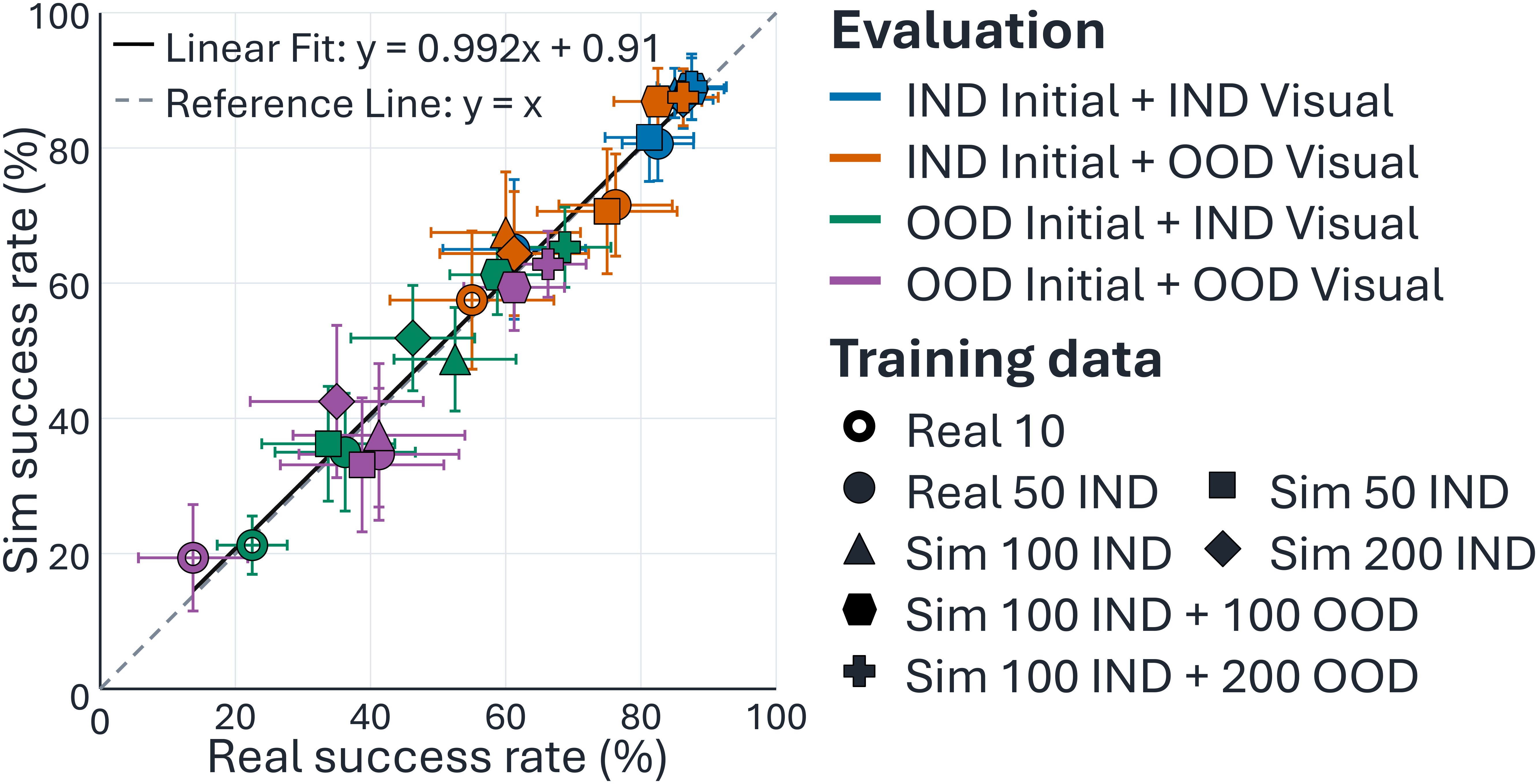}
  \vspace{-3mm}
  \caption{Sim-Real correlation of TwinScaler's policy success rates.
  Error bars indicate $\pm1$ standard error across eight tasks.}
  \vspace{-6mm}
  \label{fig:correlation_quantitative}
\end{figure}

\vspace{-2mm}
\subsection{Correlation Check}
\label{sec:correlation_check}
\vspace{-1mm}
Figure~\ref{fig:correlation_quantitative} compares 28 pairs of simulated and real-world policy success rates across seven training set configurations and four evaluation conditions. Each success rate is averaged over eight TwinScaler tasks.
Across all pairs, TwinScaler achieves a Pearson correlation of $r=0.983$ and a mean absolute error of 3.40\%.
Within individual evaluation conditions, correlations range from $0.929$ to $0.981$, with MAEs between 1.83\% and 4.91\%.
The fitted line, $y=0.992x+0.91$, lies close to the identity line $y=x$.
Overall, TwinScaler achieves sim-real consistency in policy performance trends with correlated success rates.

\vspace{-2mm}
\subsection{Ablation Studies}
\label{sec:ablation}
\vspace{-1mm}
We evaluate the contributions of optimization-based alignment and replay-based physics identification to interaction quality in simulation and downstream policy performance in the real world.
Starting from the full method, we first remove physics identification and then remove scale and pose optimization as well.
We assess interaction quality by reconstructing and replaying 160 demonstrations across 16 TwinScaler tasks in their reconstructed digital twins.
For real-world policy evaluation, we use the same eight TwinScaler tasks and four IND/OOD evaluation conditions as in Section 4.2.
For each ablation variant, we generate 100 IND and 200 OOD synthetic trajectories per task, train a $\pi_{0.5}$ policy per variant, and then measure policy success rates in the real world. Experiment results are shown in Table~\ref{tab:ablation}.

\noindent\textbf{Replay-based Physics Identification.}
The first variant retains the optimized object scales and initial poses while replacing the identified object physical parameters with VLM-estimated priors. 
Compared with the full method, this variant's replay success rate drops from 94.38\% to 78.75\%, while its real-world policy success rate drops from 77.19\% to 68.13\%.
The full method's higher replay and policy success rates highlight the importance of physics alignment in reconstruction and trajectory scaling.

\noindent\textbf{Object Scale and Pose Optimization.}
The second variant further removes object scale and pose optimization while retaining the same VLM-estimated physical priors.
It initializes object scales and translations with DA3 and object orientations with MegaPose.
As reported in Table~\ref{tab:ablation}, this variant's replay success rate drops to 57.50\%, object mask IoU decreases to 0.473, and object position error increases to 5.81\,cm, while its real-world policy success rate falls to 61.88\%.
This performance drop highlights the importance of geometric alignment in reproducing observed interactions and generating effective synthetic training data.

\noindent\textbf{Held-Out Replay for Physics Verification.}
To evaluate whether the identified object physical parameters generalize to new demonstrations of the same tasks, we collect 5 additional episodes per task and reuse the previously reconstructed digital twins. 
For each held-out episode, we only reconstruct the first-frame object states and replay the recorded robot actions, keeping the previously reconstructed assets, identified controllers and physics fixed.
Across the 80 held-out episodes, TwinScaler achieves 93.75\% replay success, 0.689 mask IoU, and 2.18\,cm position error, similar to the full method's results in Table~\ref{tab:ablation}.

\begin{table}[t]
\centering
\footnotesize
\setlength{\tabcolsep}{1pt}
\begin{tabular*}{\columnwidth}{@{\extracolsep{\fill}}lcccc@{}}
\toprule
\multirow{2}{*}{\textbf{Method}}
& \multicolumn{3}{c}{\textbf{Simulation}}
& \multicolumn{1}{c@{}}{\textbf{Real-World}} \\
\cmidrule(lr){2-4}\cmidrule(l){5-5}
& \textbf{Replay SR}$\uparrow$
& \textbf{Mask IoU}$\uparrow$
& \textbf{Pos. Err.}$\downarrow$
& \textbf{Policy SR}$\uparrow$ \\

\midrule
\shortstack[l]{w/o Scale \& Pose Opt.} & 57.50\% & 0.473 & 5.81\,cm & 61.88\% \\
w/o Physics ID. & 78.75\% & 0.669 & 3.24\,cm & 68.13\% \\
\textbf{TwinScaler Full} & \textbf{94.38\%} & \textbf{0.694} & \textbf{2.21\,cm} & \textbf{77.19\%} \\
\bottomrule
\end{tabular*}
\vspace{-3.5mm}
\caption{Sequential ablation of the replay-based object physics identification and iterative object scale and pose optimization. Both variants use the same VLM-estimated physical priors.
}
\vspace{-7mm}
\label{tab:ablation}
\end{table}

\vspace{-3mm}
\section{Conclusion}
\label{sec:conclusion}
\vspace{-2mm}

TwinScaler combines agentic scene and task understanding, optimization-based alignment, and replay-based physics identification to automate the digital twin reconstruction, identification, and trajectory scaling process from real-world robot demonstrations without additional scans.
Scaling 10 real demonstrations to 300 synthetic trajectories per task raises mean policy success rates from 40.78\% to 76.17\% in simulation and from 38.13\% to 77.19\% in the real world, with correlated sim-real policy success rates.

{\small
\bibliographystyle{ieeenat_fullname}
\bibliography{main}

@misc{Re3Sim,
      title={Re$^3$Sim: Generating High-Fidelity Simulation Data via 3D-Photorealistic Real-to-Sim for Robotic Manipulation}, 
      author={Xiaoshen Han and Junqiu Yu and Minghuan Liu and Yilun Chen and Xiaoyang Lyu and Yang Tian and Bolun Wang and Weinan Zhang and Jiangmiao Pang},
      year={2026},
      eprint={2502.08645},
      archivePrefix={arXiv},
      primaryClass={cs.RO},
      url={https://arxiv.org/abs/2502.08645}, 
}

@misc{DREAM,
      title={DREAM: Deployment-Time Demonstration Generation via Real-to-Sim for Scalable Policy Adaptation}, 
      author={Makoto Sato and Tatsuya Matsushima and Yutaka Matsuo and Yusuke Iwasawa},
      year={2026},
      eprint={2608.29078},
      archivePrefix={arXiv},
      primaryClass={cs.RO},
      url={https://arxiv.org/abs/2608.29078}, 
}

@misc{EmbodiRSI,
      title={EmbodiRSI: Recursive Self-Improvement for Data-Efficient Robot Adaptation}, 
      author={Haoran Lang and Haotao Lu and Shiyu Sang and Haoyang Luo and Guo Chen and Qun Li and Jingyi Yu and Ye Shi and Jingya Wang},
      year={2026},
      eprint={2609.38905},
      archivePrefix={arXiv},
      primaryClass={cs.RO},
      url={https://arxiv.org/abs/2609.38905}, 
}

@misc{ARSTAG,
      title={ARSTAG: An Agentic Real2Sim2Real System for Task-Specific Robot Data Generation}, 
      author={Bowei Li and Yuner Zhang and Changliu Liu},
      year={2026},
      eprint={2609.24563},
      archivePrefix={arXiv},
      primaryClass={cs.RO},
      url={https://arxiv.org/abs/2609.24563}, 
}

@misc{RoLA,
      title={Robot Learning from Any Images}, 
      author={Siheng Zhao and Jiageng Mao and Wei Chow and Zeyu Shangguan and Tianheng Shi and Rong Xue and Yuxi Zheng and Yijia Weng and Yang You and Daniel Seita and Leonidas Guibas and Sergey Zakharov and Vitor Guizilini and Yue Wang},
      year={2025},
      eprint={2509.22970},
      archivePrefix={arXiv},
      primaryClass={cs.RO},
      url={https://arxiv.org/abs/2509.22970}, 
}

@misc{SimFoundry,
      title={SimFoundry: Modular and Automated Scene Generation for Policy Learning and Evaluation}, 
      author={Nadun Ranawaka and Josiah Wong and Wei-Lin Pai and Wei-Teng Chu and Tianyuan Dai and Masoud Moghani and Hang Yin and Yunfan Jiang and Wesley Durbano and Brandon Huynh and Yu Fang and Danfei Xu and Ruohan Zhang and Li Fei-Fei and Linxi Fan and Bowen Wen and Ajay Mandlekar and Yuke Zhu},
      year={2026},
      eprint={2606.28276},
      archivePrefix={arXiv},
      primaryClass={cs.RO},
      url={https://arxiv.org/abs/2606.28276}, 
}

@misc{GenieSim,
      title={Genie Sim 3.0 : A High-Fidelity Comprehensive Simulation Platform for Humanoid Robot}, 
      author={Chenghao Yin and Da Huang and Di Yang and Jichao Wang and Nanshu Zhao and Chen Xu and Wenjun Sun and Linjie Hou and Zhijun Li and Junhui Wu and Zhaobo Liu and Zhen Xiao and Sheng Zhang and Lei Bao and Rui Feng and Zhenquan Pang and Jiayu Li and Qian Wang and Maoqing Yao},
      year={2026},
      eprint={2601.02078},
      archivePrefix={arXiv},
      primaryClass={cs.RO},
      url={https://arxiv.org/abs/2601.02078}, 
}

@misc{X-Sim,
      title={X-Sim: Cross-Embodiment Learning via Real-to-Sim-to-Real}, 
      author={Prithwish Dan and Kushal Kedia and Angela Chao and Edward Weiyi Duan and Maximus Adrian Pace and Wei-Chiu Ma and Sanjiban Choudhury},
      year={2025},
      eprint={2505.07096},
      archivePrefix={arXiv},
      primaryClass={cs.RO},
      url={https://arxiv.org/abs/2505.07096}, 
}

@misc{Real2Sem2Real,
      title={Real2Sem2Real: 3D Semantic Augmentations for Robot Learning from a Single Human RGB Demonstration}, 
      author={Matthew Strong and Ankush K. Dhawan and Julien Poffet and Giuse Pham and Jinho So and Monroe Kennedy III},
      year={2026},
      eprint={2610.28880},
      archivePrefix={arXiv},
      primaryClass={cs.RO},
      url={https://arxiv.org/abs/2610.28880}, 
}

@misc{Real2Gen,
      title={Scaling Single Human Demonstrations for Imitation Learning using Generative Foundational Models}, 
      author={Nick Heppert and Minh Quang Nguyen and Abhinav Valada},
      year={2026},
      eprint={2602.12734},
      archivePrefix={arXiv},
      primaryClass={cs.RO},
      url={https://arxiv.org/abs/2602.12734}, 
}

@misc{Video2Policy,
      title={Video2Policy: Scaling up Manipulation Tasks in Simulation through Internet Videos}, 
      author={Weirui Ye and Fangchen Liu and Zheng Ding and Yang Gao and Oleh Rybkin and Pieter Abbeel},
      year={2025},
      eprint={2502.09886},
      archivePrefix={arXiv},
      primaryClass={cs.RO},
      url={https://arxiv.org/abs/2502.09886}, 
}

@misc{Video2Sim2Real,
      title={Video2Sim2Real: Full-Stack Autonomous Dexterous Skill Acquisition from a Single Human Video}, 
      author={Yunhai Han and Jianuo Qiu and Linhao Bai and Ziyu Xiao and Zihang Zeng and Yangcen Liu and Zhaodong Yang and Shalin Jain and Wenrui Ma and Jiaqi Fu and Yuqian Zheng and Manisha Natarajan and Muhammad Zubair Irshad and Kenneth Shaw and Matthew Gombolay and Zsolt Kira and Harish Ravichandar},
      year={2026},
      eprint={2606.08828},
      archivePrefix={arXiv},
      primaryClass={cs.RO},
      url={https://arxiv.org/abs/2606.08828}, 
}

@misc{PhysTwin,
      title={PhysTwin: Physics-Informed Reconstruction and Simulation of Deformable Objects from Videos}, 
      author={Hanxiao Jiang and Hao-Yu Hsu and Kaifeng Zhang and Hsin-Ni Yu and Shenlong Wang and Yunzhu Li},
      year={2025},
      eprint={2503.17973},
      archivePrefix={arXiv},
      primaryClass={cs.CV},
      url={https://arxiv.org/abs/2503.17973}, 
}

@misc{DreMa,
      title={Dream to Manipulate: Compositional World Models Empowering Robot Imitation Learning with Imagination}, 
      author={Leonardo Barcellona and Andrii Zadaianchuk and Davide Allegro and Samuele Papa and Stefano Ghidoni and Efstratios Gavves},
      year={2025},
      eprint={2412.14957},
      archivePrefix={arXiv},
      primaryClass={cs.RO},
      url={https://arxiv.org/abs/2412.14957}, 
}

@misc{D-REX,
      title={D-REX: Differentiable Real-to-Sim-to-Real Engine for Learning Dexterous Grasping}, 
      author={Haozhe Lou and Mingtong Zhang and Haoran Geng and Hanyang Zhou and Sicheng He and Zhiyuan Gao and Siheng Zhao and Jiageng Mao and Pieter Abbeel and Jitendra Malik and Daniel Seita and Yue Wang},
      year={2026},
      eprint={2603.01151},
      archivePrefix={arXiv},
      primaryClass={cs.RO},
      url={https://arxiv.org/abs/2603.01151}, 
}

@misc{TwinAligner,
      title={TwinAligner: Visual-Dynamic Alignment Empowers Physics-aware Real2Sim2Real for Robotic Manipulation}, 
      author={Hongwei Fan and Hang Dai and Jiyao Zhang and Jinzhou Li and Qiyang Yan and Yujie Zhao and Mingju Gao and Jinghang Wu and Hao Tang and Hao Dong},
      year={2025},
      eprint={2512.19390},
      archivePrefix={arXiv},
      primaryClass={cs.RO},
      url={https://arxiv.org/abs/2512.19390}, 
}

@misc{RoboSnap,
      title={RoboSnap: One-Shot Real-to-Sim Scene Generation for Generalizable Robot Learning and Evaluation}, 
      author={Shujie Zhang and Jingkun Yi and Weipeng Zhong and Zirui Zhou and Yangkun Zhu and Hanqing Wang and Xudong Xu and Weinan Zhang and Chunhua Shen},
      year={2026},
      eprint={2607.06699},
      archivePrefix={arXiv},
      primaryClass={cs.RO},
      url={https://arxiv.org/abs/2607.06699}, 
}

@misc{AgenticReal2Sim,
      title={Agentic Real2Sim: Physics-based World Modeling with Vision-Language Agents}, 
      author={Guanxiong Chen and Qianjun Xia and Jiawei Peng and Heng Zhang and Pengyu Jing and Bole Ma and Justin Qian and Yixian Cheng and Ziyi Jiao and Bingyang Zhou and Yiduo Qu and Luoxin Ye and Kaifeng Zhang and Kunyi Wang and Weijia Zeng and Yunuo Chen and Pengzhi Yang and Ziqiu Zeng and Siyuan Luo and Huamin Wang and Chao Liu and Alan Yuille and Fan Shi and Changxi Zheng and Yunzhu Li and Chenfanfu Jiang and Peter Yichen Chen},
      year={2026},
      eprint={2607.19190},
      archivePrefix={arXiv},
      primaryClass={cs.RO},
      url={https://arxiv.org/abs/2607.19190}, 
}

@misc{Real2Gym,
      title={Real2Gym: Building Gyms from Videos, Bringing Skills to Robots}, 
      author={Kerui Ren and Yingxiang Xu and Kaiwen Song and Linning Xu and Bo Dai and Mulin Yu and Tao Lu},
      year={2026},
      eprint={2609.37089},
      archivePrefix={arXiv},
      primaryClass={cs.CV},
      url={https://arxiv.org/abs/2609.37089}, 
}

@misc{RoboSimStudio,
      title={RoboSimStudio: An Interactive Toolkit for Human and Agent Rapid Real2Sim of Robot Manipulation Tasks}, 
      author={Kaiyuan Chen and Shuangyu Xie and Uksang Yoo and Xinyue Zhu and Cael Magner and Alvin Zhu and Daehwa Kim and Eric Heiden and Erwin Coumans and Miles Macklin and Spencer Huang and Ken Goldberg},
      year={2026},
      eprint={2610.18835},
      archivePrefix={arXiv},
      primaryClass={cs.RO},
      url={https://arxiv.org/abs/2610.18835}, 
}

@misc{WorldMirror,
      title={WorldMirror: Agentic Real-to-Sim Generation of Interactive Worlds from In-the-Wild Robot Demonstrations}, 
      author={Tianling Xu and Weijie Wan and Honghao Su and Kailun Su and Junyuan Tang and Huichi Zhou and Haozheng Zhang and Qi Xiong and Ping Luo and Wenbo Ding and Tianxing Chen and Yue Chen},
      year={2026},
      eprint={2610.31834},
      archivePrefix={arXiv},
      primaryClass={cs.RO},
      url={https://arxiv.org/abs/2610.31834}, 
}

@misc{CLIP,
      title={Learning Transferable Visual Models From Natural Language Supervision}, 
      author={Alec Radford and Jong Wook Kim and Chris Hallacy and Aditya Ramesh and Gabriel Goh and Sandhini Agarwal and Girish Sastry and Amanda Askell and Pamela Mishkin and Jack Clark and Gretchen Krueger and Ilya Sutskever},
      year={2021},
      eprint={2103.00020},
      archivePrefix={arXiv},
      primaryClass={cs.CV},
      url={https://arxiv.org/abs/2103.00020}, 
}

@misc{GPT3,
      title={Language Models are Few-Shot Learners}, 
      author={Tom B. Brown and Benjamin Mann and Nick Ryder and Melanie Subbiah and Jared Kaplan and Prafulla Dhariwal and Arvind Neelakantan and Pranav Shyam and Girish Sastry and Amanda Askell and Sandhini Agarwal and Ariel Herbert-Voss and Gretchen Krueger and Tom Henighan and Rewon Child and Aditya Ramesh and Daniel M. Ziegler and Jeffrey Wu and Clemens Winter and Christopher Hesse and Mark Chen and Eric Sigler and Mateusz Litwin and Scott Gray and Benjamin Chess and Jack Clark and Christopher Berner and Sam McCandlish and Alec Radford and Ilya Sutskever and Dario Amodei},
      year={2020},
      eprint={2005.14165},
      archivePrefix={arXiv},
      primaryClass={cs.CL},
      url={https://arxiv.org/abs/2005.14165}, 
}

@misc{LLaMA,
      title={LLaMA: Open and Efficient Foundation Language Models}, 
      author={Hugo Touvron and Thibaut Lavril and Gautier Izacard and Xavier Martinet and Marie-Anne Lachaux and Timothée Lacroix and Baptiste Rozière and Naman Goyal and Eric Hambro and Faisal Azhar and Aurelien Rodriguez and Armand Joulin and Edouard Grave and Guillaume Lample},
      year={2023},
      eprint={2302.13971},
      archivePrefix={arXiv},
      primaryClass={cs.CL},
      url={https://arxiv.org/abs/2302.13971}, 
}

@misc{DROID,
      title={DROID: A Large-Scale In-The-Wild Robot Manipulation Dataset}, 
      author={Alexander Khazatsky and Karl Pertsch and Suraj Nair and Ashwin Balakrishna and others},
      year={2025},
      eprint={2403.12945},
      archivePrefix={arXiv},
      primaryClass={cs.RO},
      url={https://arxiv.org/abs/2403.12945}, 
}

@misc{OXE,
      title={Open X-Embodiment: Robotic Learning Datasets and RT-X Models}, 
      author = {{Open X-Embodiment Collaboration} and others},
      year={2025},
      eprint={2310.08864},
      archivePrefix={arXiv},
      primaryClass={cs.RO},
      url={https://arxiv.org/abs/2310.08864}, 
}

@misc{BridgeData,
      title={BridgeData V2: A Dataset for Robot Learning at Scale}, 
      author={Homer Walke and Kevin Black and Abraham Lee and Moo Jin Kim and Max Du and Chongyi Zheng and Tony Zhao and Philippe Hansen-Estruch and Quan Vuong and Andre He and Vivek Myers and Kuan Fang and Chelsea Finn and Sergey Levine},
      year={2024},
      eprint={2308.12952},
      archivePrefix={arXiv},
      primaryClass={cs.RO},
      url={https://arxiv.org/abs/2308.12952}, 
}

@misc{EgoDex,
      title={EgoDex: Learning Dexterous Manipulation from Large-Scale Egocentric Video}, 
      author={Ryan Hoque and Peide Huang and David J. Yoon and Mouli Sivapurapu and Jian Zhang},
      year={2026},
      eprint={2505.11709},
      archivePrefix={arXiv},
      primaryClass={cs.CV},
      url={https://arxiv.org/abs/2505.11709}, 
}

@misc{EgoMimic,
      title={EgoMimic: Scaling Imitation Learning via Egocentric Video}, 
      author={Simar Kareer and Dhruv Patel and Ryan Punamiya and Pranay Mathur and Shuo Cheng and Chen Wang and Judy Hoffman and Danfei Xu},
      year={2024},
      eprint={2410.24221},
      archivePrefix={arXiv},
      primaryClass={cs.RO},
      url={https://arxiv.org/abs/2410.24221}, 
}

@misc{Ego4D,
      title={Ego4D: Around the World in 3,000 Hours of Egocentric Video}, 
      author={Kristen Grauman and Andrew Westbury and Eugene Byrne and Zachary Chavis and others},
      year={2022},
      eprint={2110.07058},
      archivePrefix={arXiv},
      primaryClass={cs.CV},
      url={https://arxiv.org/abs/2110.07058}, 
}

@misc{MobileManiBench,
      title={MobileManiBench: Simplifying Model Verification for Mobile Manipulation}, 
      author={Wenbo Wang and Fangyun Wei and QiXiu Li and Xi Chen and Yaobo Liang and Chang Xu and Jiaolong Yang and Baining Guo},
      year={2026},
      eprint={2602.05233},
      archivePrefix={arXiv},
      primaryClass={cs.RO},
      url={https://arxiv.org/abs/2602.05233}, 
}

@misc{UniGraspTransformer,
      title={UniGraspTransformer: Simplified Policy Distillation for Scalable Dexterous Robotic Grasping}, 
      author={Wenbo Wang and Fangyun Wei and Lei Zhou and Xi Chen and Lin Luo and Xiaohan Yi and Yizhong Zhang and Yaobo Liang and Chang Xu and Yan Lu and Jiaolong Yang and Baining Guo},
      year={2025},
      eprint={2412.02699},
      archivePrefix={arXiv},
      primaryClass={cs.RO},
      url={https://arxiv.org/abs/2412.02699}, 
}

@misc{RoboTwin,
      title={RoboTwin 2.0: A Scalable Data Generator and Benchmark with Strong Domain Randomization for Robust Bimanual Robotic Manipulation}, 
      author={Tianxing Chen and Zanxin Chen and Baijun Chen and Zijian Cai and Yibin Liu and Zixuan Li and Qiwei Liang and Xianliang Lin and Yiheng Ge and Zhenyu Gu and Weiliang Deng and Yubin Guo and Tian Nian and Xuanbing Xie and Qiangyu Chen and Kailun Su and Tianling Xu and Guodong Liu and Mengkang Hu and Huan-ang Gao and Kaixuan Wang and Zhixuan Liang and Yusen Qin and Xiaokang Yang and Ping Luo and Yao Mu},
      year={2025},
      eprint={2506.18088},
      archivePrefix={arXiv},
      primaryClass={cs.RO},
      url={https://arxiv.org/abs/2506.18088}, 
}

@misc{MimicGen,
      title={MimicGen: A Data Generation System for Scalable Robot Learning using Human Demonstrations}, 
      author={Ajay Mandlekar and Soroush Nasiriany and Bowen Wen and Iretiayo Akinola and Yashraj Narang and Linxi Fan and Yuke Zhu and Dieter Fox},
      year={2023},
      eprint={2310.17596},
      archivePrefix={arXiv},
      primaryClass={cs.RO},
      url={https://arxiv.org/abs/2310.17596}, 
}

@misc{SkillMimicGen,
      title={SkillMimicGen: Automated Demonstration Generation for Efficient Skill Learning and Deployment}, 
      author={Caelan Garrett and Ajay Mandlekar and Bowen Wen and Dieter Fox},
      year={2024},
      eprint={2410.18907},
      archivePrefix={arXiv},
      primaryClass={cs.RO},
      url={https://arxiv.org/abs/2410.18907}, 
}

@misc{GPT6Sol,
  author       = {{OpenAI}},
  title        = {{GPT-6.1 Sol}},
  year         = {2026},
  howpublished = {\url{https://developers.openai.com/api/docs/models/gpt-6.1-sol}},
  note         = {Accessed: 2026-11-11}
}

@misc{ManiSkill,
      title={ManiSkill3: GPU Parallelized Robotics Simulation and Rendering for Generalizable Embodied AI}, 
      author={Stone Tao and Fanbo Xiang and Arth Shukla and Yuzhe Qin and Xander Hinrichsen and Xiaodi Yuan and Chen Bao and Xinsong Lin and Yulin Liu and Tse-kai Chan and Yuan Gao and Xuanlin Li and Tongzhou Mu and Nan Xiao and Arnav Gurha and Viswesh Nagaswamy Rajesh and Yong Woo Choi and Yen-Ru Chen and Zhiao Huang and Roberto Calandra and Rui Chen and Shan Luo and Hao Su},
      year={2025},
      eprint={2410.00425},
      archivePrefix={arXiv},
      primaryClass={cs.RO},
      url={https://arxiv.org/abs/2410.00425}, 
}

@misc{IsaacSim,
  author       = {{NVIDIA}},
  title        = {{Isaac Sim}},
  howpublished = {\url{https://github.com/isaac-sim/IsaacSim}},
  note         = {Accessed: 2026-11-11}
}

@misc{IsaacLab,
      title={Isaac Lab: A GPU-Accelerated Simulation Framework for Multi-Modal Robot Learning}, 
      author={NVIDIA and Mayank Mittal and Pascal Roth and James Tigue and Antoine Richard and others},
      year={2025},
      eprint={2511.04831},
      archivePrefix={arXiv},
      primaryClass={cs.RO},
      url={https://arxiv.org/abs/2511.04831}, 
}

@INPROCEEDINGS{MuJoCo,
  author={Todorov, Emanuel and Erez, Tom and Tassa, Yuval},
  booktitle={2012 IEEE/RSJ International Conference on Intelligent Robots and Systems}, 
  title={MuJoCo: A physics engine for model-based control}, 
  year={2012},
  volume={},
  number={},
  pages={5026-5033},
  doi={10.1109/IROS.2012.6386109}}

@misc{SAM,
      title={SAM 3: Segment Anything with Concepts}, 
      author={Nicolas Carion and Laura Gustafson and Yuan-Ting Hu and Shoubhik Debnath and Ronghang Hu and Didac Suris and Chaitanya Ryali and Kalyan Vasudev Alwala and Haitham Khedr and Andrew Huang and Jie Lei and Tengyu Ma and Baishan Guo and Arpit Kalla and Markus Marks and Joseph Greer and Meng Wang and Peize Sun and Roman Rädle and Triantafyllos Afouras and Effrosyni Mavroudi and Katherine Xu and Tsung-Han Wu and Yu Zhou and Liliane Momeni and Rishi Hazra and Shuangrui Ding and Sagar Vaze and Francois Porcher and Feng Li and Siyuan Li and Aishwarya Kamath and Ho Kei Cheng and Piotr Dollár and Nikhila Ravi and Kate Saenko and Pengchuan Zhang and Christoph Feichtenhofer},
      year={2026},
      eprint={2511.16719},
      archivePrefix={arXiv},
      primaryClass={cs.CV},
      url={https://arxiv.org/abs/2511.16719}, 
}

@misc{MV-SAM3D,
      title={MV-SAM3D: Adaptive Multi-View Fusion for Layout-Aware 3D Generation}, 
      author={Baicheng Li and Dong Wu and Jun Li and Shunkai Zhou and Zecui Zeng and Lusong Li and Hongbin Zha},
      year={2026},
      eprint={2603.11633},
      archivePrefix={arXiv},
      primaryClass={cs.CV},
      url={https://arxiv.org/abs/2603.11633}, 
}

@misc{Particulate,
      title={Particulate: Feed-Forward 3D Object Articulation}, 
      author={Ruining Li and Yuxin Yao and Chuanxia Zheng and Christian Rupprecht and Joan Lasenby and Shangzhe Wu and Andrea Vedaldi},
      year={2026},
      eprint={2512.11798},
      archivePrefix={arXiv},
      primaryClass={cs.CV},
      url={https://arxiv.org/abs/2512.11798}, 
}

@misc{DA3,
      title={Depth Anything 3: Recovering the Visual Space from Any Views}, 
      author={Haotong Lin and Sili Chen and Junhao Liew and Donny Y. Chen and Zhenyu Li and Guang Shi and Jiashi Feng and Bingyi Kang},
      year={2025},
      eprint={2511.10647},
      archivePrefix={arXiv},
      primaryClass={cs.CV},
      url={https://arxiv.org/abs/2511.10647}, 
}

@misc{MegaPose,
      title={MegaPose: 6D Pose Estimation of Novel Objects via Render \& Compare}, 
      author={Yann Labbé and Lucas Manuelli and Arsalan Mousavian and Stephen Tyree and Stan Birchfield and Jonathan Tremblay and Justin Carpentier and Mathieu Aubry and Dieter Fox and Josef Sivic},
      year={2022},
      eprint={2212.06870},
      archivePrefix={arXiv},
      primaryClass={cs.CV},
      url={https://arxiv.org/abs/2212.06870}, 
}

@misc{AlignPose,
      title={AlignPose: Generalizable 6D Pose Estimation via Multi-view Feature-metric Alignment}, 
      author={Anna Šárová Mikeštíková and Médéric Fourmy and Martin Cífka and Josef Sivic and Vladimir Petrik},
      year={2026},
      eprint={2512.20538},
      archivePrefix={arXiv},
      primaryClass={cs.CV},
      url={https://arxiv.org/abs/2512.20538}, 
}

@misc{3DGS,
      title={3D Gaussian Splatting for Real-Time Radiance Field Rendering}, 
      author={Bernhard Kerbl and Georgios Kopanas and Thomas Leimkühler and George Drettakis},
      year={2023},
      eprint={2308.04079},
      archivePrefix={arXiv},
      primaryClass={cs.GR},
      url={https://arxiv.org/abs/2308.04079}, 
}

@inproceedings{FEM,
    author = {Sifakis, Eftychios and Barbic, Jernej},
    title = {FEM simulation of 3D deformable solids: a practitioner's guide to theory, discretization and model reduction},
    year = {2012},
    isbn = {9781450316781},
    publisher = {Association for Computing Machinery},
    address = {New York, NY, USA},
    url = {https://doi.org/10.1145/2343483.2343501},
    doi = {10.1145/2343483.2343501},
    booktitle = {ACM SIGGRAPH 2012 Courses},
    articleno = {20},
    numpages = {50},
    location = {Los Angeles, California},
    series = {SIGGRAPH '12}
}

@misc{Qwen,
      title={Qwen-Image Technical Report}, 
      author={Chenfei Wu and Jiahao Li and Jingren Zhou and Junyang Lin and Kaiyuan Gao and Kun Yan and Sheng-ming Yin and Shuai Bai and Xiao Xu and Yilei Chen and Yuxiang Chen and Zecheng Tang and Zekai Zhang and Zhengyi Wang and An Yang and Bowen Yu and Chen Cheng and Dayiheng Liu and Deqing Li and Hang Zhang and Hao Meng and Hu Wei and Jingyuan Ni and Kai Chen and Kuan Cao and Liang Peng and Lin Qu and Minggang Wu and Peng Wang and Shuting Yu and Tingkun Wen and Wensen Feng and Xiaoxiao Xu and Yi Wang and Yichang Zhang and Yongqiang Zhu and Yujia Wu and Yuxuan Cai and Zenan Liu},
      year={2025},
      eprint={2508.02324},
      archivePrefix={arXiv},
      primaryClass={cs.CV},
      url={https://arxiv.org/abs/2508.02324}, 
}

@phdthesis{M3T,
  author = {Stoiber, Manuel},
  title  = {Closing the Loop: {3D} Object Tracking
            for Advanced Robotic Manipulation},
  school = {Technical University of Munich},
  year   = {2023},
  url    = {https://mediatum.ub.tum.de/1696259}
}

@INPROCEEDINGS{PSO,
  author={Kennedy, James and Eberhart, Russell},
  booktitle={Proceedings of ICNN'95 - International Conference on Neural Networks}, 
  title={Particle swarm optimization}, 
  year={1995},
  volume={4},
  number={},
  pages={1942--1948},
  doi={10.1109/ICNN.1995.488968}}

@misc{cuRobo,
      title={cuRobo: Parallelized Collision-Free Minimum-Jerk Robot Motion Generation}, 
      author={Balakumar Sundaralingam and Siva Kumar Sastry Hari and Adam Fishman and Caelan Garrett and Karl Van Wyk and Valts Blukis and Alexander Millane and Helen Oleynikova and Ankur Handa and Fabio Ramos and Nathan Ratliff and Dieter Fox},
      year={2023},
      eprint={2310.17274},
      archivePrefix={arXiv},
      primaryClass={cs.RO},
      url={https://arxiv.org/abs/2310.17274}, 
}

@misc{GSO,
      title={Google Scanned Objects: A High-Quality Dataset of 3D Scanned Household Items}, 
      author={Laura Downs and Anthony Francis and Nate Koenig and Brandon Kinman and Ryan Hickman and Krista Reymann and Thomas B. McHugh and Vincent Vanhoucke},
      year={2022},
      eprint={2204.11918},
      archivePrefix={arXiv},
      primaryClass={cs.RO},
      url={https://arxiv.org/abs/2204.11918}, 
}

@misc{BC,
      title={What Matters in Learning from Offline Human Demonstrations for Robot Manipulation}, 
      author={Ajay Mandlekar and Danfei Xu and Josiah Wong and Soroush Nasiriany and Chen Wang and Rohun Kulkarni and Li Fei-Fei and Silvio Savarese and Yuke Zhu and Roberto Martín-Martín},
      year={2021},
      eprint={2108.03298},
      archivePrefix={arXiv},
      primaryClass={cs.RO},
      url={https://arxiv.org/abs/2108.03298}, 
}

@misc{DiffusionPolicy,
      title={Diffusion Policy: Visuomotor Policy Learning via Action Diffusion}, 
      author={Cheng Chi and Zhenjia Xu and Siyuan Feng and Eric Cousineau and Yilun Du and Benjamin Burchfiel and Russ Tedrake and Shuran Song},
      year={2024},
      eprint={2303.04137},
      archivePrefix={arXiv},
      primaryClass={cs.RO},
      url={https://arxiv.org/abs/2303.04137}, 
}

@misc{OpenVLA,
      title={OpenVLA: An Open-Source Vision-Language-Action Model}, 
      author={Moo Jin Kim and Karl Pertsch and Siddharth Karamcheti and Ted Xiao and Ashwin Balakrishna and Suraj Nair and Rafael Rafailov and Ethan Foster and Grace Lam and Pannag Sanketi and Quan Vuong and Thomas Kollar and Benjamin Burchfiel and Russ Tedrake and Dorsa Sadigh and Sergey Levine and Percy Liang and Chelsea Finn},
      year={2024},
      eprint={2406.09246},
      archivePrefix={arXiv},
      primaryClass={cs.RO},
      url={https://arxiv.org/abs/2406.09246}, 
}

@misc{pi05,
      title={$\pi_{0.5}$: a Vision-Language-Action Model with Open-World Generalization}, 
      author={{Physical Intelligence} and Kevin Black and Noah Brown and James Darpinian and Karan Dhabalia and Danny Driess and Adnan Esmail and Michael Equi and Chelsea Finn and Niccolo Fusai and Manuel Y. Galliker and Dibya Ghosh and Lachy Groom and Karol Hausman and Brian Ichter and Szymon Jakubczak and Tim Jones and Liyiming Ke and Devin LeBlanc and Sergey Levine and Adrian Li-Bell and Mohith Mothukuri and Suraj Nair and Karl Pertsch and Allen Z. Ren and Lucy Xiaoyang Shi and Laura Smith and Jost Tobias Springenberg and Kyle Stachowicz and James Tanner and Quan Vuong and Homer Walke and Anna Walling and Haohuan Wang and Lili Yu and Ury Zhilinsky},
      year={2025},
      eprint={2504.16054},
      archivePrefix={arXiv},
      primaryClass={cs.LG},
      url={https://arxiv.org/abs/2504.16054}, 
}

@misc{WorldVLA,
      title={WorldVLA: Towards Autoregressive Action World Model}, 
      author={Jun Cen and Chaohui Yu and Hangjie Yuan and Yuming Jiang and Siteng Huang and Jiayan Guo and Xin Li and Yibing Song and Hao Luo and Fan Wang and Deli Zhao and Hao Chen},
      year={2025},
      eprint={2506.21539},
      archivePrefix={arXiv},
      primaryClass={cs.RO},
      url={https://arxiv.org/abs/2506.21539}, 
}

@misc{DreamZero,
      title={World Action Models are Zero-shot Policies}, 
      author={Seonghyeon Ye and Yunhao Ge and Kaiyuan Zheng and Shenyuan Gao and Sihyun Yu and George Kurian and Suneel Indupuru and You Liang Tan and Chuning Zhu and Jiannan Xiang and Ayaan Malik and Kyungmin Lee and William Liang and Nadun Ranawaka and Jiasheng Gu and Yinzhen Xu and Guanzhi Wang and Fengyuan Hu and Avnish Narayan and Johan Bjorck and Jing Wang and Gwanghyun Kim and Dantong Niu and Ruijie Zheng and Yuqi Xie and Jimmy Wu and Qi Wang and Ryan Julian and Danfei Xu and Yilun Du and Yevgen Chebotar and Scott Reed and Jan Kautz and Yuke Zhu and Linxi "Jim" Fan and Joel Jang},
      year={2026},
      eprint={2602.15922},
      archivePrefix={arXiv},
      primaryClass={cs.RO},
      url={https://arxiv.org/abs/2602.15922}, 
}

@misc{GenieG2,
  author       = {{AGIBOT}},
  title        = {{AGIBOT G2}},
  howpublished = {\url{https://agibot.com/products/G2}},
  note         = {Accessed: 2026-11-11}
}

@misc{MetaQuest,
  author       = {{Meta}},
  title        = {{Meta Quest}},
  howpublished = {\url{https://www.meta.com/quest/}},
  note         = {Accessed: 2026-11-11}
}

@misc{RialTo,
      title={Reconciling Reality through Simulation: A Real-to-Sim-to-Real Approach for Robust Manipulation}, 
      author={Marcel Torne and Anthony Simeonov and Zechu Li and April Chan and Tao Chen and Abhishek Gupta and Pulkit Agrawal},
      year={2024},
      eprint={2403.03949},
      archivePrefix={arXiv},
      primaryClass={cs.RO},
      url={https://arxiv.org/abs/2403.03949}, 
}

@misc{ACDC,
      title={Automated Creation of Digital Cousins for Robust Policy Learning}, 
      author={Tianyuan Dai and Josiah Wong and Yunfan Jiang and Chen Wang and Cem Gokmen and Ruohan Zhang and Jiajun Wu and Li Fei-Fei},
      year={2024},
      eprint={2410.07408},
      archivePrefix={arXiv},
      primaryClass={cs.RO},
      url={https://arxiv.org/abs/2410.07408}, 
}

@misc{SIMPLER,
      title={Evaluating Real-World Robot Manipulation Policies in Simulation}, 
      author={Xuanlin Li and Kyle Hsu and Jiayuan Gu and Karl Pertsch and Oier Mees and Homer Rich Walke and Chuyuan Fu and Ishikaa Lunawat and Isabel Sieh and Sean Kirmani and Sergey Levine and Jiajun Wu and Chelsea Finn and Hao Su and Quan Vuong and Ted Xiao},
      year={2024},
      eprint={2405.05941},
      archivePrefix={arXiv},
      primaryClass={cs.RO},
      url={https://arxiv.org/abs/2405.05941}, 
}

@misc{PolaRiS,
      title={PolaRiS: Scalable Real-to-Sim Evaluations for Generalist Robot Policies}, 
      author={Arhan Jain and Mingtong Zhang and Kanav Arora and William Chen and Marcel Torne and Muhammad Zubair Irshad and Sergey Zakharov and Yue Wang and Sergey Levine and Chelsea Finn and Wei-Chiu Ma and Dhruv Shah and Abhishek Gupta and Karl Pertsch},
      year={2025},
      eprint={2512.16881},
      archivePrefix={arXiv},
      primaryClass={cs.RO},
      url={https://arxiv.org/abs/2512.16881}, 
}

@misc{LPIPS,
      title={The Unreasonable Effectiveness of Deep Features as a Perceptual Metric}, 
      author={Richard Zhang and Phillip Isola and Alexei A. Efros and Eli Shechtman and Oliver Wang},
      year={2018},
      eprint={1801.03924},
      archivePrefix={arXiv},
      primaryClass={cs.CV},
      url={https://arxiv.org/abs/1801.03924}, 
}

@ARTICLE{SSIM,
  author={Zhou Wang and Bovik, A.C. and Sheikh, H.R. and Simoncelli, E.P.},
  journal={IEEE Transactions on Image Processing}, 
  title={Image quality assessment: from error visibility to structural similarity}, 
  year={2004},
  volume={13},
  number={4},
  pages={600-612},
  doi={10.1109/TIP.2003.819861}}
}

\end{document}